\def\year{2019}\relax
\documentclass[letterpaper]{article} %DO NOT CHANGE THIS
\usepackage{aaai}  %Required
\usepackage{times}  %Required
\usepackage{helvet}  %Required
\usepackage{courier}  %Required
\usepackage{url}  %Required
\usepackage{graphicx}  %Required
\usepackage{sidecap}
\usepackage{latexsym}
\usepackage{url}
\usepackage{CJKutf8}
\usepackage{amsmath}
\usepackage{multirow}
\usepackage{amssymb}
\usepackage{booktabs}
\usepackage[disable]{todonotes}
\usepackage{verbatim}
\usepackage{caption}
\usepackage{subcaption}

\newcommand{\newcite}[1]{\citeauthor{#1} [\citeyear{#1}]}

\newcommand{\Note}[2]{} 
\newcommand{\SideNote}[2]{} 
\renewcommand{\Note}[2]{\todo[color=#1,size=\small, inline=true]{#2}} 
\renewcommand{\SideNote}[2]{\todo[color=#1,size=\small]{#2}} % 

\makeatletter
\renewcommand{\paragraph}{%
  \@startsection{paragraph}{4}%
  {\z@}{.9ex \@plus 1ex \@minus .2ex}{-1em}%
  {\normalfont\normalsize\bfseries}%
}
\makeatother
\begin{document}
\title{Plan-And-Write: Towards Better Automatic Storytelling}
\author{Lili Yao$^{1,3}$\thanks{Equal contribution: Lili Yao and Nanyun Peng}, Nanyun Peng$^{2*}$, Ralph Weischedel$^2$, Kevin Knight$^2$, Dongyan Zhao$^1$ and Rui Yan$^1$\thanks{Corresponding author: Rui Yan (ruiyan@pku.edu.cn)} \\
\texttt{liliyao@tencent.com, \{npeng,weisched,knight\}@isi.edu}\\
\texttt{\{zhaodongyan,ruiyan\}@pku.edu.cn}\\
$^1$ Institute of Computer Science and Technology, Peking University\\
$^2$ Information Sciences Institute, University of Southern California, 
$^3$ Tencent AI Lab
}

\begin{CJK*}{UTF8}{gkai}
\maketitle

%\begin{comment}

\begin{abstract}
  Automatic storytelling is challenging since it requires generating long,  coherent natural language to describes a sensible sequence of events. Despite considerable efforts on automatic story generation in the past, prior work either is restricted in plot planning, or can only generate stories in a narrow domain. 
In this paper, we explore open-domain story generation that writes stories given a title (topic) as input. We propose a {\em plan-and-write} hierarchical generation framework that first plans a storyline, and then generates a story based on the storyline. We compare two planning strategies. The {\em dynamic} schema interweaves story planning and its surface realization in text, while the {\em static} schema plans out the entire storyline before generating stories. 
Experiments show that with explicit storyline planning, the generated stories are more diverse, coherent, and on topic than those generated without creating a full plan, according to both automatic and human evaluations. 
\end{abstract}

\section{Introduction}

A narrative or story is anything which is told in the form of a causally/logically linked set of events involving some shared characters \cite{mostafazadeh-EtAl:2016}.
Automatic storytelling requires composing coherent natural language texts that describe a sensible sequence of events. This seems much harder than text generation where a plan or knowledge fragment already exists. Thus, story generation seems an ideal testbed for advances in general AI. 
Prior research on story generation mostly focused on automatically composing a sequence of events that can be told as a story by plot planning~\cite{lebowitz1987planning,perez2001mexica,porteous2009controlling,riedl2010narrative,li2013story} %porteous2010applying,
or case-based reasoning~\cite{turner1994creative,gervas2005story}. 
These approaches rely heavily on human annotation and/or are restricted to limited domains. 
Moreover, most prior work is restricted to the abstract story representation level without surface realization in natural language.

In this paper, we study generating natural language stories from any given title (topic). 
Inspired by prior work on dialog planning \cite{nayak2017plan} and narrative planning \cite{riedl2010narrative}, we propose to decompose story generation into two steps: 1) story planning which generates plots, and 2) surface realization which composes natural language text based on the plots. 
We propose a {\it plan-and-write} hierarchical generation framework that combines plot planning and surface realization to generate stories from titles.  

\begin{table}[t]    
  \centering
  \resizebox{\linewidth}{!}{%
    \begin{tabular}{p{2cm}|p{5.3cm}}
      \toprule[1pt] %\hline
      \bf{Title (Given)} & The Bike Accident\\ \hline
      \bf{Storyline (Extracted)} & Carrie $\to$ bike $\to$ sneak $\to$ nervous $\to$ leg \\ \hline
      \bf{Story \newline (Human Written)} & \underline{Carrie} had just learned how to ride a bike. She didn't have a \underline{bike} of her own. Carrie would \underline{sneak} rides on her sister's bike. She got \underline{nervous} on a hill and crashed into a wall. The bike frame bent and Carrie got a deep gash on her \underline{leg}. \\
    \bottomrule[1pt]
    \end{tabular}%
  }
  \caption{An example of title, storyline and story in our system. A storyline is represented by an ordered list of words.}
  \label{introcase}
\vspace{-.5cm}
\end{table}

One major challenge for our framework is how to represent and obtain annotations for story plots so that a reasonable generative model can be trained to plan story plots. 
\newcite{li2013story} introduces plot graphs which contain events and their relations to represent a storyline. 
Plot graphs are comprehensive representations of story plots, however, the definition and curation of such plot graphs require highly specialized knowledge and significant human effort. 
On the other hand, in poetry composition, \newcite{wang2016chinese} provides a sequence of words to guide poetry generation. 
In conversational systems, \newcite{mou2016sequence} takes keywords as the main gist of the reply to guide response generation. 
We take a similar approach to represent a story plot with a sequence of words. Specifically, we use the order that the words appear in the story to approximate a storyline. 
Table~\ref{introcase} shows an example of the title, storyline, and story. 

Though this representation seems to over-simplify story plots, it has several advantages. 
First, because the storyline representation is simple, there are many reliable tools to extract high-quality storylines from existing stories and thus automatically generate training data for the plot planning model. % without relying on manual annotations. 
Our experiments show that by training plot planning models on automatically extracted storylines, we can generate better stories without additional human annotation. 
Moreover, with this simple and interpretable storyline representation, it is possible to compare the efficiency of different plan-and-write strategies. 
Specifically, 
we explore two paradigms that seem to mimic human practice in real world story writing\footnote{Some discussions on Quora: \url{https://www.quora.com/} \url{How-many-times-does-a-writer-edit-his-first} \url{-draft}}~\cite{alarcon2010secret}. The \textit{dynamic} schema adjusts the plot improvisationally while writing progresses. The \textit{static} schema plans the entire plot before writing. 
We summarize the contributions of the paper as follows:
\begin{itemize}
\item We propose a {\em plan-and-write} framework that leverages storylines to improve the diversity and coherence of the generated story. 
Two strategies: \textit{dynamic} and \textit{static} planning are explored and compared under this framework. 
\item We develop evaluation metrics to measure the diversity of the generated stories, and conduct novel analysis to examine the importance of different aspects of stories for human evaluation. 
\item Experiments show that the proposed plan-and-write model generates more diverse, coherent, and on-topic stories than those without planning~\footnote{Code and appendix will be available at \url{https://bitbucket.org/VioletPeng/language-model}}.
\end{itemize}

\begin{figure*}[tb]
  \centering
  \includegraphics[width=.8\linewidth]{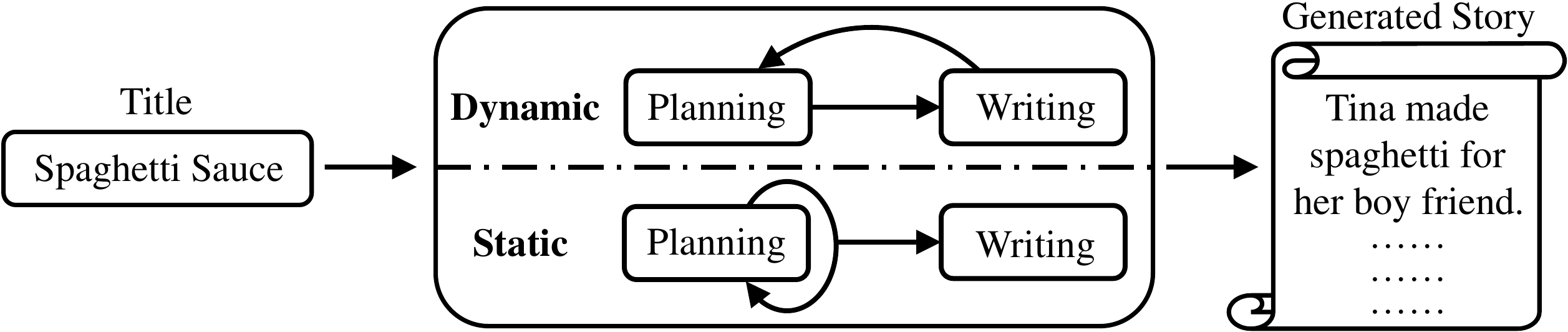}
  \vspace{-.2cm}
  \caption{An overview of our system.}. 
  \label{framework}
\vspace{-.8cm}
\end{figure*}

\section{Plan-and-Write Storytelling}
In this paper, we propose a plan-and-write framework to generate stories from given titles. We posit that storytelling systems can benefit from storyline planning to generate more coherent and on-topic stories. 
An additional benefit of the plan-and-write schema is that human and computer can interact and collaborate on the (abstract) storyline level, which can enable many potentially enjoyable interactions. 
We formally define the input, output, and storyline of our approach as follows. 

\subsection{Problem Formulation} 
\quad \textbf{Input:} A title $\mathbf{t}=\{t_1,t_2,...,t_n\}$ is given to the system to constrain writing, where $t_i$ is the $i$-th word in the title.

\textbf{Output:} The system generates a story $\mathbf{s}=\{s_1,s_2,...,s_m\}$ based on a title, where $s_i$ denotes a sentence in the story.

\textbf{Storyline:} The system plans a storyline $\mathbf{l} = \{l_1, l_2,...,l_m\}$ as an intermediate step to represent the plot of a story. We use a sequence of words to represent a storyline, therefore, $l_i$ denotes a word in a storyline.  

Given a title, the plan-and-write framework always plans a storyline. %generate coherent and on-topic stories. 
We explore two variations of this framework: the {\em dynamic} and the static schema. 

\subsection{Storyline Preparation}
\label{sec:storyline}
To obtain training data for the storyline planner, we extract sequences of words from existing story corpora to compose storylines. 
Specifically, we extract one word from each sentence of a story to form a storyline\footnote{For this pilot study, we assume each word $l_i$ in a storyline corresponds to a sentence $s_i$ in a story.}. 
We adopt the RAKE algorithm \cite{rose2010automatic}, which combines several word frequency based and graph-based metrics to 
weight the importance of the words. We extract the most important word from each sentence as a story's storyline. 

\section{Methods}
We adopt neural generation models to implement our plan-and-write framework, as they have been shown effective in many text generation tasks such as machine translation~\cite{bahdanau2014neural}, and dialogue systems~\cite{shang2015neural}. 
Figure~\ref{framework} demonstrates the workflow of our framework. We now describe the two plan-and-write strategies we explored. 

\begin{figure}
  \centering
  \begin{subfigure}[b]{.38\textwidth}
  \includegraphics[width=\textwidth]{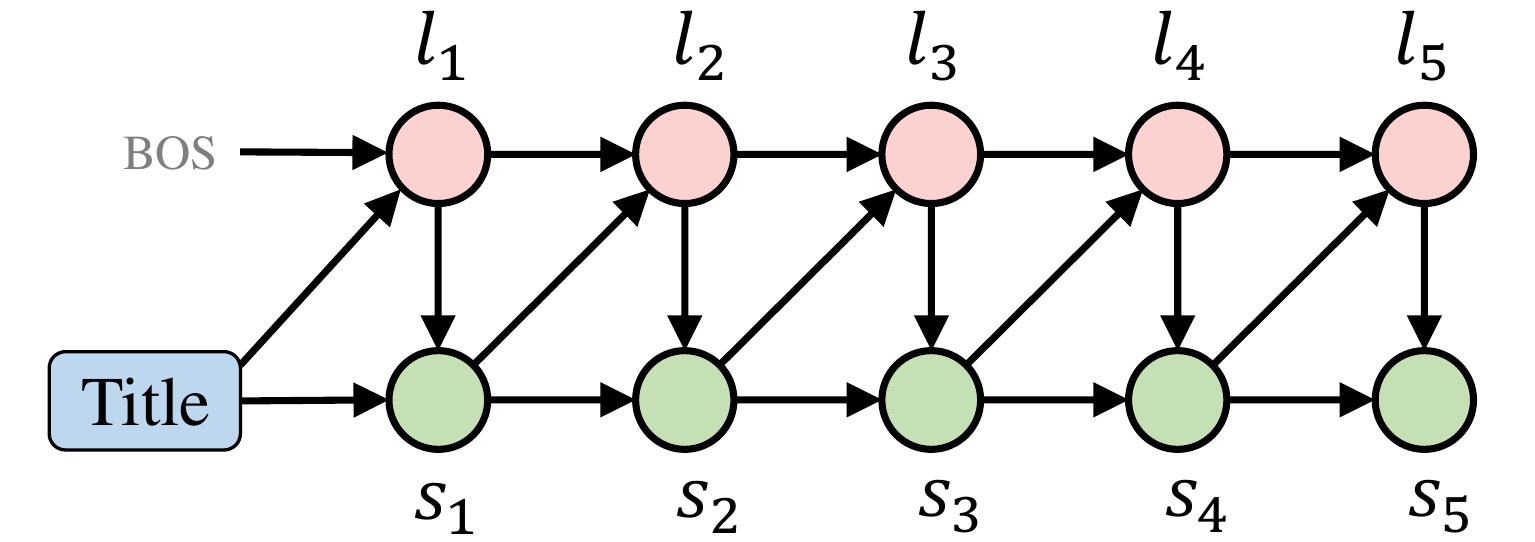}
  \caption{Dynamic schema work-flow.}
  \label{dynamic}
\end{subfigure}
  \begin{subfigure}[b]{.38\textwidth}
  \includegraphics[width=\textwidth]{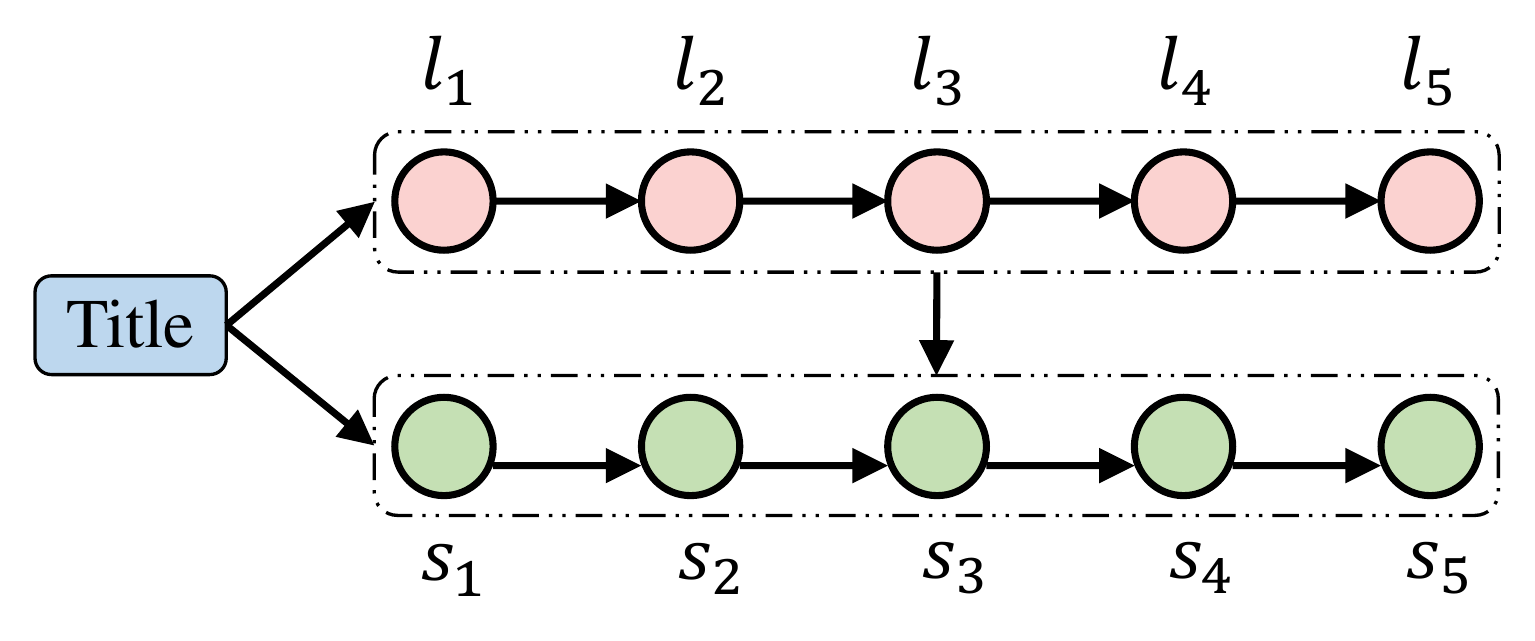}
  \caption{Static schema work-flow.} 
  \label{static}
\end{subfigure}
\caption{An illustration of the dynamic and static  plan-and-write work-flow. $l_i$ denotes a {\em word} in a storyline and $s_i$ denotes a {\em sentence} in a story.}
\vspace{-.5cm}
\end{figure}

\subsection{Dynamic Schema}
\label{sec:dynamic}
The dynamic schema emphasizes flexibility. As shown in Figure~\ref{dynamic}, it generates the next word in the storyline and the next sentence in the story at each step. 
In both cases, the existing storyline and previously generated sentences are given to the model to move one step forward. 

\subsubsection{Storyline Planning} 
The storyline is planned out based on the context (the title and previously generated sentences are taken as context) and the previous word in the storyline. 
We formulate it as a content-introducing generation problem, where the new content (the next word in the storyline) is generated based on the context and some additional information (the most recent word in the storyline). 
Formally, let $\mathbf{ctx} = [\mathbf{t,s}_{1:i-1}]$ denotes the context, where $s_{1:i-1}$ denotes for the first $i$-1 sentences in the story.  We model 
$p(l_i|\mathbf{ctx}, l_{i-1}; \theta).$ 
We implement the content-introducing method proposed by~\newcite{yao2017towards}, which first encodes context into a vector using a bidirectional gated recurrent unit (BiGRU), and then incorporates the auxiliary information, in this case the previous word in the storyline, into the decoding process. Formally, hidden vectors for context are computed as 
$\mathbf{\widetilde{h}}_{ctx} = Encode_{ctx}(\mathbf{ctx}) = 
[\overrightarrow{\mathbf{h}_{ctx}}; \overleftarrow{\mathbf{h}_{ctx}}],$
where $\overrightarrow{\mathbf{h}_{ctx}}$ and $\overleftarrow{\mathbf{h}_{ctx}}$ are the hidden vectors produced by a forward and a backward GRU, respectively. $[;]$ denotes element-wise concatenation. The conditional probability is computed as: 
\begin{align}
\small
& h_y = \textsc{GRU}(\textsc{BOS}, C_{att}),
 h_w = \textsc{GRU}(l_{i-1}, C_{att}) \nonumber \\
& h_y^{'} = \tanh (W_1 h_{y}), h_w^{'} = \tanh (W_2 h_{w}) \nonumber \\
& k = \sigma(W_k [h_y^{'};h_w^{'}]) \nonumber \\
& p(l_i|\mathbf{ctx}, l_{i-1}) = g(k \circ h_y + (1-k) \circ h_w) \nonumber
\end{align} 
\textsc{BOS} denotes the beginning of decoding. $C_{att}$ represents the attention-based context computed from  $\mathbf{\tilde{h}}_{ctx}$. g($\cdot$) denotes a multilayer perceptron (MLP).

\subsubsection{Story Generation}
The story is generated incrementally by planning and writing alternately. We formulate it as another content-introducing generation problem which generates a story sentence based on both the context and an additional storyline word as a cue. The model structure is exactly the same as for storyline generation. However, there are two differences between storyline and story generation. On one hand, the former aims to generate a word while the latter generates a variable-length sequence. On the other hand, the auxiliary information they use is different. 

Formally, the model is trained to minimize the negative log-probability of training data: 
\begin{align}
\small
\mathcal{L}(\theta)_{dyna}=\frac{1}{N} \sum_{j=1}^N \left[-\log\prod_{i=1}^m p(s_{i}|\mathbf{ctx}, l_{i})\right]_j
\end{align}
where $N$ is the number of stories in training data; $m$ denotes the number of sentences in a story.
Given the extracted storylines as described in the previous Section, the storyline and story generation models are trained separately. End-to-end generation is conducted in a pipeline fashion.

\subsection{Static Schema}
The static schema is inspired by sketches that writers usually draw before they flesh out the whole story. As illustrated in Figure~\ref{static}, it first generates a whole storyline which does not change during story writing. 
This sacrifices some flexibility in writing, but could potentially enhance story coherence as it provides ``look ahead'' for what happens next. 

\subsubsection{Storyline Planning}
Differing from the dynamic schema, storyline planning for static schema is solely based on the title $\mathbf{t}$. We formulate it as a conditional generation problem, where the probability of generating each word in a storyline depends on the previous words in the storyline and the title. Formally, we model $p(l_i|\mathbf{t}, l_{1:i-1};\theta)$. 
We adopt a sequence-to-sequence (Seq2Seq), conditional generation model that first encodes the title into a vector using a bidirectional long short-term memory network (BiLSTM), and generates words in the storyline using another single-directional LSTM.
Formally, the hidden vector $\mathbf{\tilde{h}}$ for a title is computed as 
$\mathbf{\tilde{h}} = Encode(\mathbf{t}) = 
[\overrightarrow{\mathbf{h}}; \overleftarrow{\mathbf{h}}],
$ and the conditional probability is given by:
\begin{align}
\small
p(l_i|\mathbf{t}, l_{1:i-1};\theta)  = g(\textsc{LSTM}_{\text{att}}(\mathbf{\tilde{h}}, l_{i-1}, \mathbf{\mathbf{h}}^{\text{dec}}_{i-1})) \nonumber
\end{align}
where $\textsc{LSTM}_{\text{att}}$ denotes a cell of the LSTM with attention mechanism \cite{bahdanau2014neural}; $\mathbf{\mathbf{h}}^{\text{dec}}_{i-1}$ stands for the decoding hidden state. g($\cdot$) again denotes a MLP.

\subsubsection{Story Generation}
The story is generated after the full storyline is planned. We formulate it as another conditional generation problem. 
Specifically, we train a Seq2Seq model that encodes both the title and the planned storyline into a low-dimensional vector by first concatenating them with a special symbol $<$EOT$>$ in between, and encode them with BiLSTMs:
$
\mathbf{\tilde{h}}_{tl} = Encode_{tl}(\mathbf{[t,l]}) = 
[\overrightarrow{\mathbf{h}_{tl}}; \overleftarrow{\mathbf{h}_{tl}}].
$
The Seq2Seq model is then trained to minimize the negative log-probability of the stories in the training data:
\begin{align}
\small
\mathcal{L}(\theta)_{static}=\frac{1}{N} \sum_{j=1}^N \left[-\log\prod_{i=1}^m p(s_i|\mathbf{\tilde{h}}_{tl}, s_{1:i-1})\right]_j
\end{align}

\subsection{Storyline Optimization}
\label{sec:storyline-opt}
One common problem for neural generation models is the repetition in generated results~\cite{li2016deep}. We observe repetition initially in both the generated storyline (repeated words) and story (repeated phrases and sentences). 
An advantage of the storyline layer is that given the compact and interpretable representation of the storyline, we can easily apply heuristics to reduce repetition\footnote{It is important to avoid repetition in the generated stories too. However, it is hard to automatically detect repetition in stories. Optimizing storylines can indirectly reduce repetition in stories.}. Specifically, we forbid any word to appear twice when generating a storyline.

\begin{table}[t]
  \centering
    \begin{tabular}{cc} \hline
    \toprule[0.5pt]
      Number of Stories & $98,161$ \\
      Vocabulary size & $33,215$ \\
      Average number of words & $50$ \\ 
    \bottomrule[0.5pt]
    \end{tabular}%
  \caption{Statistics of the ROCStories dataset.}
  \label{statistics}   
  \vspace{-.5cm}
\end{table}

\section{Experimental Setup}
\label{experiment-setup}
\subsection{Dataset}
We conduct the experiments on the ROCStories corpus~\cite{mostafazadeh-EtAl:2016}. It contains 98,162 short commonsense stories as training data, and additional 1,817 stories for development and test, respectively. The stories in the corpus are five-sentence stories that capture a rich set of causal and temporal commonsense relations between daily events, making them a good resource for training storytelling models. Table~\ref{statistics} shows the statistics of ROCStories dataset. 
Since only the training set of the ROCStories corpus contains titles, which we need as input. We split the original training data into 8:1:1 for training, validation, and testing. 

\subsection{Baselines}
To evaluate the effectiveness of the plan-and-write framework, we compare our methods against representative baselines without a planning module.

\noindent \paragraph{Inc-S2S} denotes the incremental sentence-to-sentence generation baseline, which creates stories by generating the first sentence from a given title, then generating the $i$-th sentence from the title and the previously generated $i$-1 sentences. This resembles the \textit{dynamic} schema without planning. We use a Seq2Seq model with attention ~\cite{bahdanau2014neural} to implement the Inc-S2S baseline, where the sequence to sequence model is trained to generate the next sentence based on the context.

\noindent \paragraph{Cond-LM} denotes the conditional language model baseline, which straightforwardly generates the whole story word by word from a given title. Again we use a Seq2Seq model with attention as our implementation of the conditional language model, where the sequence to sequence model is trained to generate the whole story based on the title. It resembles our \textit{static} schema without planning.
\begin{figure*}[t!]
  \centering
  \includegraphics[width=0.96\textwidth]{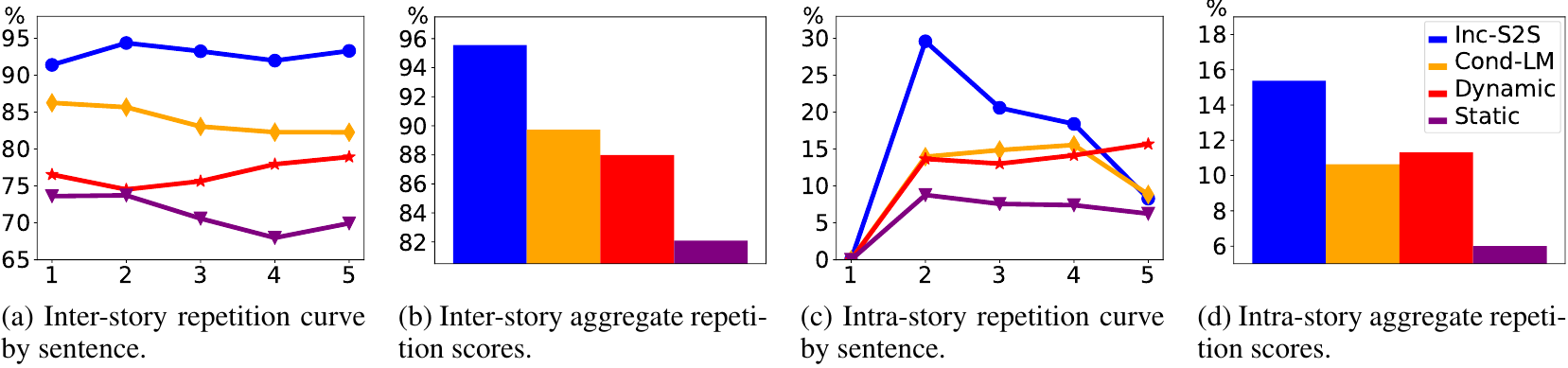}
    \caption{Inter- and intra-story repetition rates by sentences (curves) and for the whole stories (bars), the lower the better. As reference points, the aggregate repetition rates on the human-written training data are 34\% and 0.3\% for the inter- and intra-story measurements respectively.}
  \label{repetition}
  \vspace{-.5em}
\end{figure*}

\begin{comment}
\begin{figure*}[t!]
  \centering
  \begin{subfigure}[t]{0.22\textwidth}
  \includegraphics[width=\textwidth]{interline3.pdf}
  \caption{Inter-story repetition curve by sentence.}
  \end{subfigure}
  \hspace{.5em} 
  \begin{subfigure}[t]{0.223\textwidth}  
  \includegraphics[width=\textwidth]{interbar3.pdf}
  \caption{Inter-story aggregate repetition scores.}
  \end{subfigure}
  \hspace{.5em} 
  \begin{subfigure}[t]{0.22\textwidth}
  \includegraphics[width=\textwidth]{intraline3.pdf}
  \caption{Intra-story repetition curve by sentence. }
  \end{subfigure}
  \hspace{.5em}
  \begin{subfigure}[t]{0.223\textwidth}
  \includegraphics[width=\textwidth]{intrabar3.pdf}
  \caption{Intra-story aggregate repetition scores.}
  \end{subfigure}
  \caption{Inter- and intra-story repetition rates by sentences (curves) and for the whole stories (bars), the lower the better. As reference points, the aggregate repetition rates on the human-written training data are 34\% and 0.3\% for the inter- and intra-story measurements respectively.}
  \label{repetition}
  \vspace{-.5em}
\end{figure*}

\end{comment}

\begin{table*}[h!]
  \centering
  %\resizebox{\textwidth}{!}{ %
    \begin{tabular}{c|c|c|c||c|c|c||c|c|c} \hline
    \toprule[1.5pt] \hline
      \multirow{2}{*}{Choice \%} & \multicolumn{3}{c||}{Dynamic \textit{vs} Inc-S2S} & \multicolumn{3}{c||}{Static \textit{vs} Cond-LM} & \multicolumn{3}{c}{Dynamic \textit{vs} Static} \\ \cline{2-10}
      & Dyna. & Inc. & Kappa & Static & Cond. & Kappa & Dyna. & Static & Kappa \\ \hline 
%      Seq2Seq & 2.135 & 0.622 & 0.255 & * & * & * \\
     Fidelity & \bf35.8  &  12.9 & 0.42 & \bf 38.5  & 16.3 & 0.42 & 21.47 & \bf38.00 & 0.30 \\ 
     Coherence & \bf 37.2 & 28.6 & 0.30 & \bf 39.4 & 32.3 & 0.35 & 28.27 & \bf49.47 & 0.36 \\
     Interestingness & \bf 43.5 & 26.7 & 0.31 & \bf 39.5 & 35.7 & 0.42 & 34.40 & \bf42.60 & 0.35\\ 
     Overall Popularity & \bf 42.9 & 27.0 & 0.34 & \bf 40.9 & 34.2 & 0.38 & 30.07 & \bf50.07 & 0.38\\ \hline
    \bottomrule[1.5pt]
    \end{tabular}%
% } 
  \caption{Human evaluation results on four aspects: fidelity, coherence, interestingness, and overall user preference. Dyna., Inc., and Cond. is the abbreviation for Dynamic schema, Inc-S2S, and Cond-LM respectively. We also calculate the Kappa coefficient to show the inter-annotator agreement.}
  \vspace{-.3cm}
  \label{human-eval}   
\end{table*}

\subsection{Hyper-parameters}
As all of our baselines and the proposed methods are RNN-based conditional generation models, we conduct the same set of hyper-parameter optimization for them. We train all the models using stochastic gradient descent (SGD). For the encoder and decoder in our generation models, we tune the hyper-parameters of the embedding and hidden vector dimensions and the dropout rate by grid search. 
We randomly initialize the word embeddings and tune the dimensions in the range of [100, 200, 300, 500] for storyline generation and [300, 500, 1000] for story generation. We tune the hidden vector dimensions in the range of [300, 500, 1000]. 
The embedding and hidden vector dropout rates are all tuned from 0 to 0.5, step by 0.1. 
We tune all baselines and proposed models based on BLEU scores~\cite{papineni2002bleu} on the validation set. Details of the best hyper-parameter values for each setting are given in Appendix.

\subsection{Evaluation Metrics}
\label{metric}
\subsubsection{Objective metrics.} 
Our goal is generating human-like stories, which can pass the Turing test. Therefore, the evaluation metrics based on n-gram overlap such as BLEU are not suitable for our task\footnote{Our plan-and-write methods also improve BLEU scores over the baseline methods, more details can be found in Appendix.}. To better gauge the quality of our methods, we design novel automatic evaluation metrics to evaluate the generation results at scale. 
Since neural generation models are known to suffer from generating repetitive content, our automatic evaluation metrics are designed to quantify diversity across the generated stories. 
We design two measurements to gauge inter- and intra-story repetition. For each sentence position $i$, the inter-story $r_{e}^i$ and intra-story $r_{a}^i$ repetition rate are computed as follows:
\begin{equation}
\small
\label{single-rep}
\begin{aligned}
r_{e}^i &= 1 - \frac{T(\sum_{j=1}^N s^{ji})}{T_{all}(\sum_{j=1}^N  s^{ji})} \\
r_{a}^i &= \frac{1}{N}\sum_{j=1}^N \left[ \frac{\sum_{k=1}^{i-1}T(s^i\cap s^{k})}{(i-1) * T(s^i)}\right]^j
\end{aligned}
\end{equation}
where $T(\cdot)$ and $T_{all}(\cdot)$ denote the number of distinct and total trigrams\footnote{We also conduct the same computation for four and five-grams and observed the same trends. The Spearman correlation between this measurement and human rating is 0.28.}, respectively. $s^{ji}$ stands for the $i$-th sentence in $j$-th story; $s^i\cap s^{k}$ is the distinct trigram intersection set between sentence $s^i$ and $s^k$. Naturally, $r_{e}^i$ demonstrates the repetition rate between stories at sentence position $i$; $r_{a}^i$ embodies the average repetition of sentence $s^i$ comparing with former sentences in a story.

We compute the aggregate scores as follows:
\begin{equation}
\small
\label{agg-rep}
\begin{aligned}
r_{e}^{agg} &= 1 - \frac{T(\sum_{j=1}^{N} \sum_{i=1}^{m} s^{ji})}{T_{all}(\sum_{j=1}^{N} \sum_{i=1}^{m} s^{ji})} \\
r_{a}^{agg} &= \frac{1}{m}\sum_{i=1}^m r_{a}^i
\end{aligned}
\end{equation}
where $\sum_{j=1}^{N} \sum_{i=1}^{m} s^{ji}$ is the set of $N$ stories with $m$ sentences. In our experiments, we set $m=5$. $r_e^{agg}$ indicates the overall repetition of all stories.

\subsubsection{Subjective metrics.} 
For a creative generation task such as story generation, reliable automatic evaluation metrics to assess aspects such as interestingness, coherence are lacking. Therefore, we rely on human evaluation to assess the quality of generation. 
We conduct pairwise comparisons, and provide users two generated stories, asking them to choose the better one. 
We consider four aspects: \textit{fidelity} (whether the story is on-topic with the given title), \textit{coherence} (whether the story is logically consistent and coherent), \textit{interestingness} (whether the story is interesting) and \textit{overall user preference} (how do users like the story). 
All surveys were collected on Amazon Mechanical Turk (AMT). 

\section{Results and Discussion}
\subsection{Objective evaluation} 
We generate 9816 stories based on the titles in the held-out test set, and compute the repetition ratio (the lower, the better) as described in Eq.~\ref{single-rep} and Eq.~\ref{agg-rep} to evaluate the diversity of the generated system. As is shown in Figure~\ref{repetition}, the proposed plan-and-write framework significantly reduces the repetition rate and generates more diverse stories. For inter-story repetition, plan-and-write methods significantly outperform all non-planning methods on individual sentences and aggregate scores. 
For the intra-story repetition rate, plan-and-write methods outperform their corresponding non-planning baselines on aggregate scores. However, the dynamic schema generates more repetitive final sentences than the baselines. 

\begin{table*}[th!]  
  \centering
%  \resizebox{\textwidth}{!}{
    \begin{tabular}{l|l|p{13.5cm}}
      \toprule[1.5pt] \hline
      \multicolumn{3}{c}{\textbf{Title: Computer}} \\ \hline % &  \textbf{A Used Car} 
       \multirow{2}{*}{\textbf{Baselines}} & Inc-S2S & Tom's computer broke down. He needed to buy a new computer. He decided to buy a new computer. Tom bought a new computer. Tom was able to buy a new computer. \\ \cline{2-3}
   & Cond-LM & The man bought a new computer. He went to the store. He bought a new computer. He bought the computer. He installed the computer. \\ \hline
      \multirow{2}{*}{\textbf{Dynamic}} & Storyline & needed $\to$ money $\to$ computer $\to$ bought  $\to$ happy \\ \cline{2-3}
       & Story & John \underline{needed} a computer for his birthday. He worked hard to earn \underline{money}. John was able to buy his \underline{computer}. He went to the store and \underline{bought} a computer. John was \underline{happy} with his new computer. 
 \\ \hline
      \multirow{2}{*}{\textbf{Static}} & Storyline & computer $\to$ slow $\to$ work $\to$ day $\to$ buy \\ \cline{2-3}
   & Story & I have an old \underline{computer}. It was very \underline{slow}. I tried to \underline{work} on it but it wouldn't work. One \underline{day}, I decided to buy a new one. I \underline{bought} a new computer . \\\hline
   \hline 
    \multicolumn{3}{c}{\textbf{Title: The Virus}} \\ \hline
       \multirow{2}{*}{\textbf{Baselines}} & Inc-S2S  & His computer was fixed and he fixed it. John got a new computer on his computer. John was able to fix it himself. John was able to fix his computer and was able to fix his computer. John was able to fix his computer and had a virus and was able to fix his computer. \\ \cline{2-3}
       & Cond-LM &  Tim was working on a project. He was working on a project. Tim was working on a project. The project was really good. Tim was able to finish the project.\\ \hline
   \multirow{2}{*}{\textbf{Dynamic}} & Storyline & computer $\to$ use $\to$ anywhere $\to$ house $\to$ found  \\ \cline{2-3}
       & Story & I was working on my \underline{computer} today. I was trying to \underline{use} the computer. I couldn't find it \underline{anywhere}. I looked all over the \underline{house} for it. Finally, I \underline{found} it.      
 \\ \hline
      \multirow{2}{*}{\textbf{Static}} & Storyline & work $\to$ fix $\to$ called $\to$ found $\to$ day \\ \cline{2-3}
       & Story & I had a virus on my computer. I tried to \underline{fix} it but it wouldn't work. I \underline{called} the repair company. They came and \underline{found} the virus. The next \underline{day}, my computer was fixed. \\ \hline
    \bottomrule[1.5pt]
    \end{tabular}%
%  }
  \caption{Case studies of generated storylines and stories.}
  \label{casestudy}
 \vspace{-.2cm}
\end{table*}

\begin{table*}[t]  
  \centering
  \begin{tabular}{p{3.3cm}|p{13.7cm}}
      \toprule[1.5pt] \hline
  \textbf{Title / Problem} &  \textbf{Story} \\ \hline
 \textbf{Taxi / off-topic} & I got a new car. It was one day. I decided to drive to the airport. I was driving for a long time. I had a great time .
\\ \hline
 \textbf{Cut / repetitive } & Anna was cutting her nails. She cut her finger and cut her finger. Then she cut her finger. It was bleeding! Anna had to bandage her finger.
 \\ \hline
 \textbf{Eight glasses/ inconsistent} & Joe needed glasses. He went to the store to buy some. He did n't have any money. He found a pair that he liked. He bought them.
 \\ \hline
\bottomrule[1.5pt]
\end{tabular}
\caption{Example stories that demonstrate the typical problems of the current systems.}
\label{analysis}
\vspace{-.5cm}
\end{table*}

\subsection{Subjective evaluation} 
For human evaluation, we randomly sample 300 titles from the test data, and present a story title and two generated stories at a time\footnote{
We compare the plan-and-write methods with their corresponding baselines and with each other. For fairness, the two stories are pooled and randomly permuted. 
Five judgments are required to reduce the variance in estimation.} to the evaluators and ask them to decide which of the two stories is better\footnote{The four aspects are each evaluated.}. There are 233 Turkers\footnote{We applied qualification filters that only allow users who have at least 500 previous jobs and had greater than 98\% acceptance rate to participate in our survey.} participated in the evaluation. Specifically, 69, 77, and 87 Turkers evaluate the comparison between Dynamic and Inc-S2S, Static and Cond-LM,  Dynamic and Static, respectively. 

Table~\ref{human-eval} demonstrates the results of human evaluation. Similar to the automatic evaluation results, both dynamic and static schema significantly outperform their counterpart baseline in all evaluation aspects, thus demonstrating the effectiveness of the proposed plan-and-write framework. Among them, the static schema shows the best results.

\begin{SCfigure}%[t]
  \centering
 \includegraphics[width=.16\textwidth]{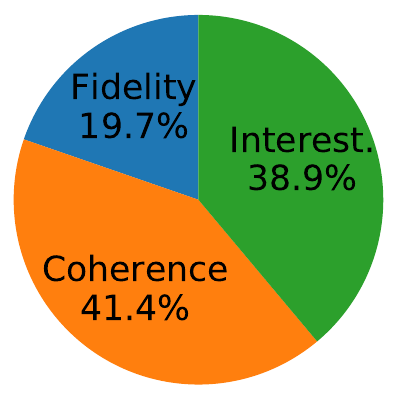}
 \caption{The regression coefficient that shows which aspect is more important in human evaluation of stories.}
 \label{regression}
 \vspace{-1em}
\end{SCfigure}

To understand why people prefer one story over another, we analyze how people weigh the three aspects (fidelity, coherence, and interestingness) in their preference for stories. We train a linear regression using the three aspects' scores as features to predict the overall score. 
We fit the regression with all human assessments we collected. 
The weight assigned to each aspect reflects their relative importance. As evident in Figure~\ref{regression}, coherence and interestingness play important roles in the human evaluation, and fidelity is less important.

\begin{table}[h]
\centering
    \begin{tabular}{c|c|c|c}
    \toprule[1.5pt] \hline
    Method & $\mathbf{l}$-B1 & $\mathbf{l}$-B2 &  $\mathbf{l}$-$\mathbf{s}$ \\ \hline
    Dynamic & 6.46 & 0.79 & 0.88 \\
    Static & 9.53 & 1.59 & 0.89 \\ \hline
    \bottomrule[1.5pt]
    \end{tabular}
\caption{The storyline BLEU score (only BLEU-1 and BLEU-2) and the correlation of storyline-story $\mathbf{l}$-$\mathbf{s}$.}
\vspace{-1em}
\label{correlation}
\end{table}

\subsection{Analysis}
The previous sections examine the overall performance of our methods quantitatively. In this section, we qualitatively analyze our methods with a focus on comparing the dynamic and static schema.

\textbf{Storyline analysis.} First, we measure the quality of the generated storylines, and the correlations between a storyline and the generated story. We use BLEU scores to measure the quality of the storylines, and an embedding-based metric~\cite{liu2016not} to estimate the average greedy matching score $\mathbf{l}$-$\mathbf{s}$ between storyline words and generated story sentences. 
%We make a small modification to not compute sentence-level embeddings; instead, 
Concretely, a storyline word is greedily matched with each token in a story sentence based on the cosine similarity of their word embeddings\footnote{For fairness, We adopt the pre-trained Glove embedding to measure the correlation. \url{http://nlp.stanford.edu/data/glove.840B.300d.zip}}. The highest cosine score is regarded as the correlation between them. 

Table~\ref{correlation} shows the results. We can see that the static schema generates storylines with higher BLEU scores.  
It also generates stories that have a higher correlation with the storylines (higher $\mathbf{l}$-$\mathbf{s}$ score\footnote{There are 75\% and 78\% storyline words appear in the generated stories in the dynamic and static schema, respectively.}). 
This indicates that with better storylines (higher BLEU score), it is easier to generate more relevant and coherent stories. This partially explains why the static schema performs better than the dynamic schema.

\textbf{Case study.} We further present two examples in Table~\ref{casestudy} to intuitively compare the plan-and-write methods and the baselines\footnote{More examples please see our live demo at \url{http://cwc-story.isi.edu/}}. 
In both examples, the baselines without planning components tend to generate repetitive sentences that do not exhibit much of a story progression. In contrast, the plan-and-write methods can generate storylines that follow a reasonable flow, and thus help generate coherent stories with less repetition. 
This demonstrates the ability of the plan-and-write methods. %However, the events are more concrete and diverse in the static schema-generated story. 
In the second example, the storyline generated by the dynamic schema is not very coherent and thus significantly affects story quality. 
This reflects the importance of storyline planning in our framework.
%However, the generated story is more creative than static strategy. 

\textbf{Error analysis.} To better understand the limitation of our best system, we manually reviewed 50 titles and the corresponding generated stories from our static schema to conduct error analysis. The three major problems are: off-topic, repetitive, and logically inconsistent. We show three examples, one for each category, in Table~\ref{analysis} to illustrate the problems. We can see that the current system is already capable of generating grammatical sentences that are coherent within a local context. However, generating a sequence of coherent and logically consistent sentences is still an open challenge.

\section{Related work}  
\subsection{Story Planning}

Automatic story generation efforts date back to the 1970s~\cite{meehan1977tale}. Early attempts focused on composing a sensible plot for a story. %relied on symbolic planning or case-based reasoning. 
Symbolic planning systems~\cite{porteous2009controlling,riedl2010narrative} %,porteous2010applying
attempted to select and sequence character actions according to specific success criteria. Case-based reasoning systems~\cite{turner1994creative,gervas2005story,Montfort06naturallanguage} adapted prior story plots (cases) to new storytelling requirements. These traditional approaches were able to produce impressive results based on hand-crafted, well-defined domain models, which kept track of legal characters, their actions, narratives, and user interest. However, the generated stories were restricted to limited domains.

To tackle the problem of restricted domains, some work attempted to automatically learn domain models. 
~\newcite{swanson2012say} mined millions of personal stories from the Web and identified relevant existing stories in the corpus. ~\newcite{li2013story} used a crowd-sourced corpus of stories to learn a domain model that helped generate stories in unknown domains. 
These efforts stayed at the level of story plot planning without surface realization.

\subsection{Event Structures for Storytelling}
There is a line of research focusing on representing story event structures \cite{mostafazadeh2016caters,mcdowell2017event}. 
\newcite{Rishes:2013} presents a model that reproduce different versions of a story from its symbolic representation. \newcite{pichotta2016learning} parse a large collection of natural language documents, extract sequences of events, and learn statistical models of them. 
Some recent work explored story generation with additional information~\cite{bowden2016m2d,peng2018towards,guan2019story}. 
Visual storytelling~\cite{huang2016visual,liu2016storytelling,wang2018no} aims to generate human-level narrative language from a sequence of images. 
\newcite{JainAMSLS17} addresses the task of coherent story generation from independent textual descriptions. Unlike this line of work, we learn to {\it automatically generate} storylines to help generate coherent stories. 

\newcite{martin2017event,xu2018skeleton} are the closest work to ours, which decomposed story generation into two steps: story structure modeling and structure-to-surface generation. 
However, \newcite{martin2017event} did not conduct experiments on full story generation. 
\newcite{xu2018skeleton} is a concurrent work which is similar to our dynamic schema. 
Their setting assumes story prompts as inputs, which is more specific than our work (which only requires a title). 
Moreover, we explore two planning strategies: dynamic schema and static schema, and show the latter works better.

\subsection{Neural Story Generation}
Recently, deep learning models have been demonstrated effective in natural language generation tasks \cite{bahdanau2014neural,merity2017regularizing}
In story generation, prior work has proposed to use deep neural networks to capture story structures and generate stories. 
\newcite{khalifa2017deeptingle} argue that stories are better generated using recurrent neural networks (RNNs) trained on highly specialized textual corpora, such as a body of work from a single, prolific author. 
\newcite{roemmele2017rnn} use skip-thought vectors~\cite{kiros2015skip} to encode sentences and model relations between the sentences. 
\newcite{JainAMSLS17} explore generating coherent stories from independent textual descriptions based on two conditional text-generation methods: statistical machine translation and sequence-to-sequence models. 
\newcite{fan2018hierarchical} proposes a hierarchical generation strategy to generate stories from prompts to improve coherence. 
%However, they assume manually curated story prompts while we learn a model to generate storylines. 
However, we consider storylines are different from prompts as they are not naturally language sentences. They are some structured outline of stories.
We employ neural network-based generation models for our plan-and-write generation. The focus, however, is to introduce storyline planning to improve the quality of generated stories, and compare the effect of different storyline planning strategies on story generation.

\section{Conclusion and Future Work}
In this paper, we propose a {\em plan-and-write} framework that generates stories from given titles with explicit storyline planning. 
We explore and compare two plan-and-write strategies: dynamic schema and static schema, and show that they both outperform the baselines without planning components. 
The static schema performs better than the dynamic schema because it plans the storyline holistically, thus tends to generate more coherent and relevant stories.

The current plan-and-write models use a sequence of words to approximate a storyline, which simplifies many meaningful structures in a real story plot. We plan to extend the exploration to richer representations, such as entity, event, and relation structures, to depict story plots. 
We also plan to extend the plan-and-write framework to generate longer documents. 
The current framework relies on storylines automatically extracted from story corpora to train the planning module. In the future, we will explore the storyline induction and joint storyline and story generation to avoid error propagation in the current pipeline generation system. 

\section*{Acknowledgements}
We thank the anonymous reviewers for the useful comments. This work is supported by Contract W911NF-15-1-0543 with the US Defense Advanced Research Projects Agency (DARPA), the National Key Research and Development Program of China (No. 2017YFC0804001), and the National Science Foundation of China (Nos. 61876196 and 61672058).
%The current plan-and-write models use a sequence of words to approximate storylines, which is a simplification as a pilot study. We plan to explore richer representations, such as event relation graphs, to represent story plots. 

%\end{comment}

\bibliography{aaai2019}
\bibliographystyle{aaai}

\clearpage
\appendix
\section{Appendix}
%\subsection{Details of implementation parameters}
\label{hyper-param} 

\begin{figure*}[th]
\centering
\includegraphics[width=\linewidth]{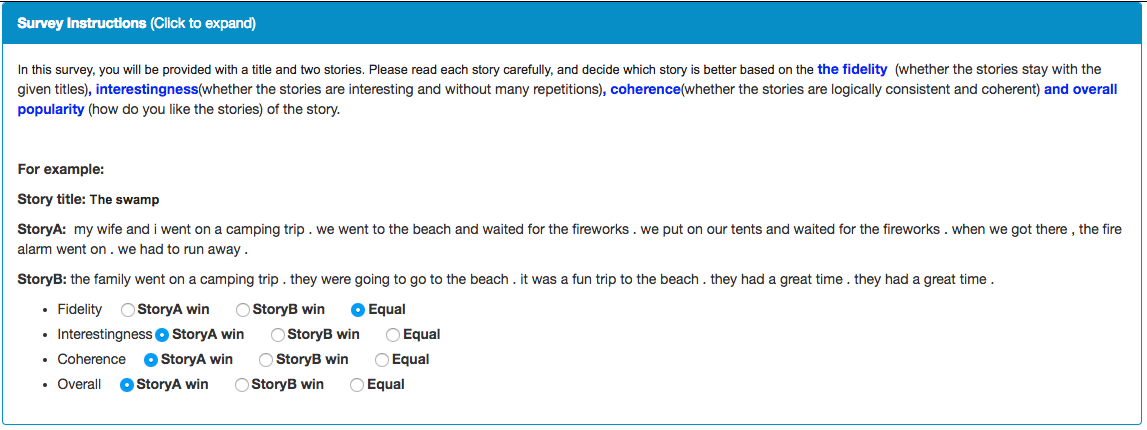}
  \caption{A snapshot of the survey on AMT for human evaluation.}. 
  \label{survey}
\end{figure*}

In this section, we describe the best hyper-parameter values for each setting in Table~\ref{hyper} and present BLEU score in Table~\ref{bleu-test}. To provide an intuitive understanding of human evaluation process, we show a snapshot of the survey on AMT in Figure~\ref{survey}.

 \begin{table}[h]
   \centering
   \resizebox{\columnwidth}{!}{%
     \begin{tabular}{c|c|c|c|c} \hline
     \toprule[1.5pt] \hline
       \textbf{Method} & E-Dim & H-Dim & E-Drop & H-Drop\\ \hline 
       \textbf{Inc-S2S} & 500  & 500  & 0.0  & 0.5 \\ 
       \textbf{Cond-LM} & 500 & 1000 & 0.4 & 0.4 \\
       \textbf{Dynamic-$\mathbf{l}$}  & 500  & 500  & 0.0 & 0.5 \\
       \textbf{Dynamic-$\mathbf{s}$}  & 500  & 500  & 0.0 & 0.5 \\
       \textbf{Static-$\mathbf{l}$}  &  500 & 1000  & 0.4 & 0.1 \\ 
       \textbf{Static-$\mathbf{s}$}  & 500  & 1000  & 0.2 & 0.1 \\ \hline
     \bottomrule[1.5pt]
     \end{tabular}%
 } 
   \caption{Best hyper-parameter settings for the baselines and our models. ``-Dim" denotes dimension, ``-Drop" represents dropout rate, and ``E" and ``H'' denote the embedding and hidden layer. ``-$\mathbf{l}$'' and ``-$\mathbf{s}$'' stand for storyline and story respectively.}
   \label{hyper}
 \end{table}

 \begin{table}[h]
  \centering
    \begin{tabular}{c|c|c|c|c} \hline
    \toprule[1.5pt] \hline
      \textbf{Method} & B-1 & B-2 & B-3 & B-4\\ \hline 
      \textbf{Inc-S2S} & 25.13  & 10.22  & 4.65 & 2.35 \\ 
      \textbf{Cond-LM} & 28.07 & 11.62 & 5.11 & 2.55 \\ \hline
      \textbf{Dynamic-W.O.}  & 26.16  & 10.39  & 4.62 & 2.32 \\
      \textbf{Static-W.O.} & 27.16 & 12.21 & 6.04 & 3.27 \\
      \textbf{Dynamic}  & 28.47 & 11.49  & 5.21  & 2.62 \\  
      \textbf{Static} & 28.20 & 12.80 & 6.36 & 3.44 \\ \hline
      \hline
    \bottomrule[1.5pt]
    \end{tabular}%

  \caption{BLEU scores in testing set. Suffix ``-W.O.'' denotes NO optimization for storylines. }
  \label{bleu-test}  
\end{table}

\end{CJK*}

\end{document}